\documentclass{article}
\usepackage[hyphens]{url}
\usepackage{amsmath,graphicx}
\usepackage{subfiles}
\usepackage{color}
\usepackage{booktabs}
\usepackage{amsfonts}
\usepackage{slashbox}
\usepackage{hyperref}
\usepackage{float}
\usepackage{multirow}
\usepackage{array}
\usepackage{subcaption}
\usepackage[dvipsnames]{xcolor}
\usepackage[preprint]{spconf}

\graphicspath{{images/}{../images/}}

\title{Domain Robust Feature Extraction for Rapid Low Resource ASR Development}
\name{Siddharth Dalmia$^*$, Xinjian Li\sthanks{\hspace{0.2cm}Equal Contribution}, Florian Metze and Alan W. Black}
\address{Language Technologies Institute, Carnegie Mellon University; Pittsburgh, PA; U.S.A.\\
\texttt{\{sdalmia|xinjianl|fmetze|awb\}@cs.cmu.edu}}

\begin{document}
\copyrightnotice{\copyright\ IEEE 2018}
                 \toappear{Published in the 2018 IEEE Workshop on Spoken Language Technology (SLT 2018), Athens, Greece}

\maketitle
\begin{abstract}


Developing a practical speech recognizer for a low resource language is challenging, not only because of the (potentially unknown) 
properties of the language, but also because test data may not be from the same domain as the available training data. 

In this paper, we focus on the latter challenge, i.e. domain mismatch, for systems trained using a sequence-based criterion. We demonstrate the effectiveness of using a pre-trained English recognizer, which is robust to such mismatched conditions, as a domain normalizing feature extractor on a low resource language. In our example, we use Turkish Conversational Speech and Broadcast News data. 


This enables rapid development of speech recognizers for new languages which can easily adapt to any domain. Testing in various cross-domain scenarios, we achieve relative improvements of around 25\,\% in phoneme error rate, with improvements being around 50\,\% for some domains.

\end{abstract}


\section{Introduction}

There is usually a lack of in-domain training data for low resource languages, but it is often easier to collect audio data from a different domain for training for eg. using scripted speech (religious audio books), broadcast news or conversation speech. This data may often not match the required domain of the original test data. Such out-of-domain audio might have different speaking styles, channels, background noise etc. Deep Neural Networks (DNN) based speech recognizers can be quite sensitive to such data mismatches and can fail completely when tested on unseen data. 

Data Augmentation has been found to be quite useful in these cases~\cite{ko2015audio}. This technique can be used to produce variations of training data that match the acoustic conditions of the evaluation data. Recently, the authors in ~\cite{peddinti2015jhu} built a robust model aimed at performing well in mismatched data conditions. They employed various techniques like data augmentation, time delay neural networks and i-vectors to deal with late reverberations, unseen channels and speaker variability. 

However, for scenarios like DARPA's LORELEI project~\cite{lorelei}, which require rapid development of ASR systems on low resource languages, these post-processing steps are a dispensable overhead, and there is instead a need for domain-normalizing features which can be extracted in advance and used cross-lingually or multi-lingually depending on the amount of data available in the language provided.



In this paper, we take an English ASR model which has been trained to perform well in various reverberation settings for multiple different domains of speech and use it to extract ``domain invariant'' features for training models in Turkish. We demonstrate that this cross-lingual transfer of features makes the model robust towards speaking style and channel variability by showing significant improvement in the ASR performance for cross-domain experiments, i.e., Broadcast News to Scripted Speech, Conversational Speech to Broadcast News.
Interestingly, we also see considerable improvements for in-domain experiments, i.e., within Broadcast News and Conversational Speech, indicating higher quality features than filter banks and the feasibility of a future ``domain invariant'' feature extractor.

We focus on techniques that can be pre-computed and be adapted to a new language rapidly. The techniques that we discuss ensure that there is minimum language-dependent engineering or domain expertise needed to develop the speech recognizer. We begin by discussing various research directions that have been explored in the fields of robustness in ASR and low resource ASR development, and how our work compares with the existing research. We establish a baseline, explain the datasets used and outline the different domain mismatch scenarios that we explored. This is followed by qualitative visualization to explain the working of the domain normalizing features. We provide a detailed discussion on the performance of the acoustic model on various mismatch conditions, along with some diagnostic experiments.

\section{Related Work and Datasets}

\subsection{Related Work}
Applying DNN-based speech recognition models on domain mismatched data has been one of the major problems in the speech recognition community. Several challenges including CHiME~\cite{barker2015third}, ASpIRE~\cite{harper2015automatic} and REVERB~\cite{kinoshita2016summary} have been proposed to facilitate creation of robust speech recognition systems under various acoustic environments. A large number of techniques have been proposed to overcome this issue, which can be largely classified into 3 categories: robust feature extraction, data augmentation and unsupervised domain adaptation.

\noindent{\bf Robust Feature Extraction:}
While most speech recognition systems depend on filterbank or MFCCs as input features, it was demonstrated that using robust features could improve accuracy over unseen environment~\cite{mitra2016coping}. Some examples of such robust features are gammatone wavelet cepstral coefficients (GWCC)~\cite{adiga2013gammatone}, normalized amplitude modulation feature~\cite{mitra2012normalized} and damped oscillator cepstral coefficient~\cite{mitra2013damped}.

In addition to applying each robust feature separately, fusing those robust features after encoding each using different CNNs was shown to improve performance significantly~\cite{mitra2016coping}. Another line of work has shown that extracting bottleneck features can be useful to increase recognition accuracy by using either denoising autoencoder~\cite{gehring2013extracting} or RBM~\cite{yu2011improved}.

\noindent{\bf Data Augmentation:}
Data augmentation is an alternative way for improving the robustness of speech recognition. Augmenting data by different approaches enables the model to be prepared for unseen audio environments. There are various strategies to perform data augmentation. For instance, one study reported that augmenting the audio with different speed factors can be helpful for the model~\cite{ko2015audio}. Adding noise or reverberation was also shown to make recognition more robust~\cite{karafiat2015three, hsiao2015robust}. 

\noindent{\bf Unsupervised Domain Adaptation:} 
In addition, there has been work on unsupervised methods for adapting the neural network model to the test data acoustics. The authors in~\cite{swietojanski2014learning,swietojanski2015differentiable} adapt their acoustic model to test data by learning an adaptation function between the hidden unit contributions of the training data and the development data. However, as pointed by~\cite{meng2017unsupervised}, these methods require reliable tri-phone alignment and this may not always be successful in a mismatched condition. 

Recently, there has been a lot of work~\cite{meng2017unsupervised,sun2017unsupervised,shinohara2016adversarial} using adversarial training to adapt to the target domain data in a completely unsupervised way. The authors in~\cite{sun2017unsupervised} and~\cite{shinohara2016adversarial} use a gradient reversal layer to train a domain classifier and try to learn domain invariant representations by passing a negative gradient on classification of domains. The use of Domain Seperation Networks~\cite{meng2017unsupervised} has also shown significant improvements by having two separate networks: a domain invariant network and a network that is unique to the domain.

Although there has been a lot of relevant work towards robustness to mismatched conditions, there has been very little work that shows its working on models trained using a sequence based criterion. Further, techniques like unsupervised domain adaptation and data augmentation either require information about the testing conditions or require presence of some test-domain data for adaptation. This makes these approaches expensive in terms of both time and expertise required, making them irrelevant for our use case. Thus, we have not compared our work to these approaches in our experiments.

\noindent{\bf Rapid Development of ASR:} 
There have been a lot of projects that need rapid adaptation and development of ASR systems with little to no data~\cite{schultz1997fast,metze2015semi,xue2014fast}. There has been works that show that using the features from the bottleneck layer of a large multilingual model~\cite{vu2012multilingual,thomas2012multilingual,knill2013investigation,grezl2014adaptation} or transferring weights from a pre-trained ASR~\cite{miao2014distributed,huang2013cross,ghoshal2013multilingual,stolcke2006cross} can improve recognition and convergence of ASR in low resource languages. Capturing speaker and environment information in the form of i-vectors~\cite{dehak2011front} has been shown to be useful for adaptation of neural networks~\cite{xue2014fast,miao2015speaker}. However, most of these methods are trained using conventional HMM/DNN systems which requires careful crafting of context dependent phonemes and can be time consuming and expensive.  

Recently,~\cite{dalmia2018sequence,tongfast} studied these behaviours in sequence based models. Authors in~\cite{tongfast} showed how phonological features can be exploited to have faster development of ASR. Authors in~\cite{dalmia2018sequence} showed that in a multi-lingual setting, it is beneficial to just train on large amounts of well-prepared data in any language. They also showed that including different languages introduced the model to various acoustic properties which helped the model generate better language-invariant features, thereby improving the cross-lingual performance.

Although most of this prior work looked at cross-lingual aspects of development of low resource ASR, the issue of domain mismatch between training and testing data has not been addressed. In this paper we look at how we can build sequence based ASR models for low resource languages where there is a domain mismatch in the data which is available for adaptation to the language.





\subsection{Datasets}
We use three datasets for this work: Turkish Conversational dataset (\textcolor{Red}{Conversational Speech}), Turkish Scripted dataset (\textcolor{Blue}{Scripted Speech}) and Turkish Broadcast News dataset (\textcolor{ForestGreen}{Broadcast News}). 

Turkish Conversational dataset and Scripted dataset are part of the Babel corpus provided by the IARPA Babel Program~(IARPA Babel Turkish Full Language Pack IARPA-babel105b-v0.5). This contains telephone conversation speech recorded at 8khz. To build \textcolor{Red}{Conversational Speech} systems, we use the standard Full Language Pack provided along-with the dataset~(around 80 hours of training data). The dataset also comes with \textcolor{Blue}{Scripted Speech} audio which is recorded by providing prompt sheets to speakers, and asking them to read some text or answer a short question. Since the Turkish Scripted Dataset is small and mismatched to the other corpora, we use it for testing only. While testing our models, we do not score on human noises and silences to allow cross-domain testing of the models.

The \textcolor{ForestGreen}{Broadcast News} corpus (LDC2012S06, LDC2012T21) has approximately 130 hours audio from Voice of America (VOA) in Turkish. The data is sampled with 16kHz originally, and it was down-sampled to 8kHz for our experiments. About 5\,\% of the data is randomly selected as the test set. This data is a public subset of the data used in~\cite{arisoy2009turkish}, so the results are not directly comparable.


\section{Pipeline}

\begin{figure*}[ht!]
    \centering
    \begin{subfigure}[t]{0.31\textwidth}
        \centering
\includegraphics[width=1.0\linewidth]{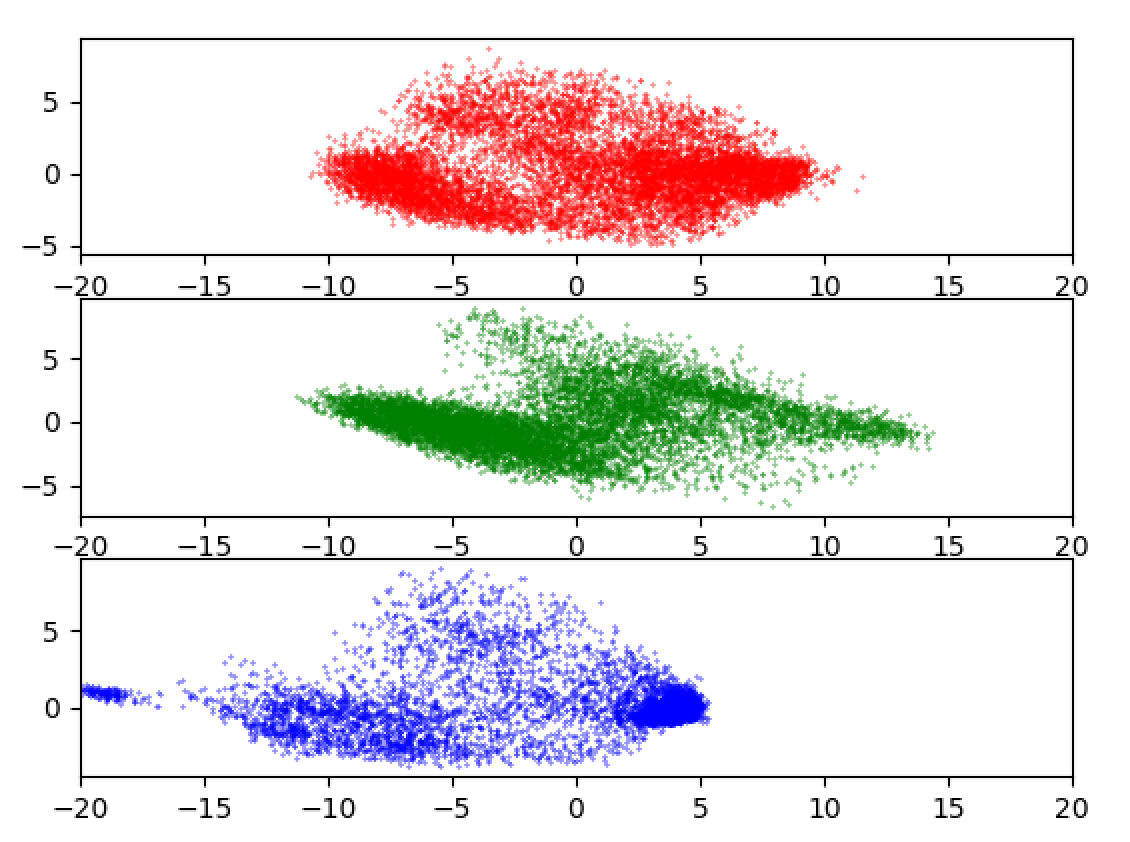} \\ \caption{Baseline Input Features}
    \end{subfigure} \hfill 
    ~ 
    \begin{subfigure}[t]{0.30\textwidth}
        \centering
\includegraphics[width=1.0\linewidth]{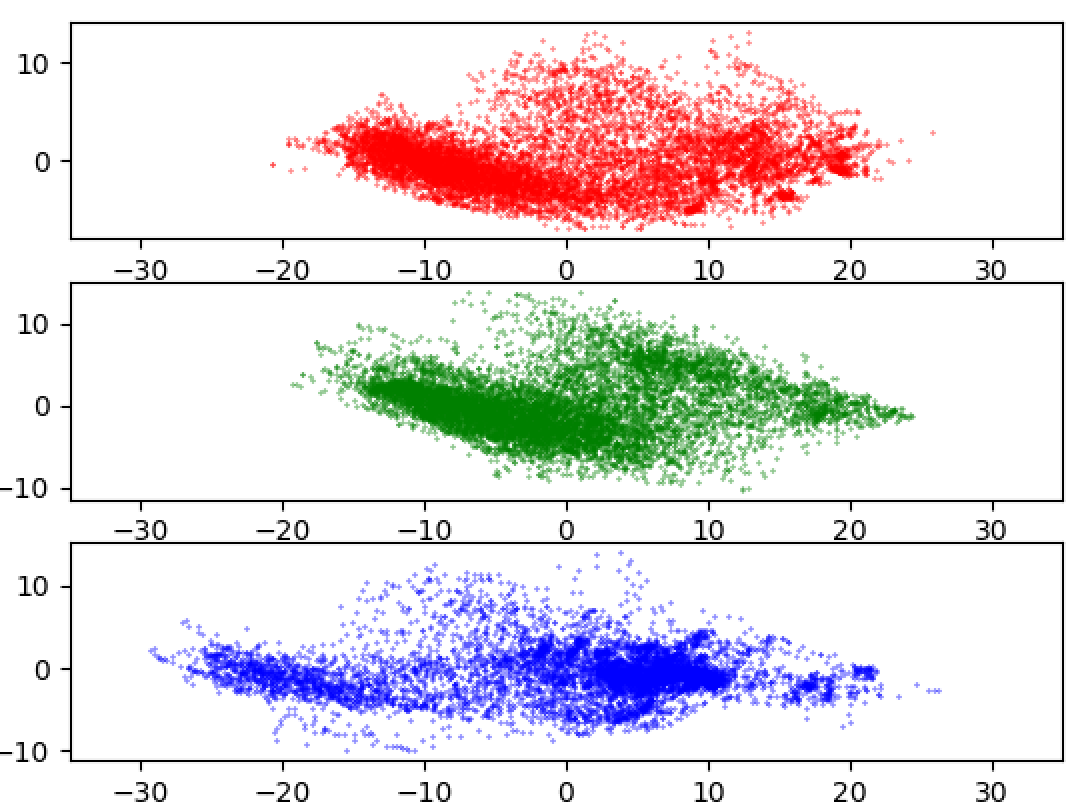}  \\   \caption{Input to ASpIRE Model}
    \end{subfigure} \hfill 
        ~ 
      \begin{subfigure}[t]{0.31\textwidth}
        \centering
\includegraphics[width=1.0\linewidth]{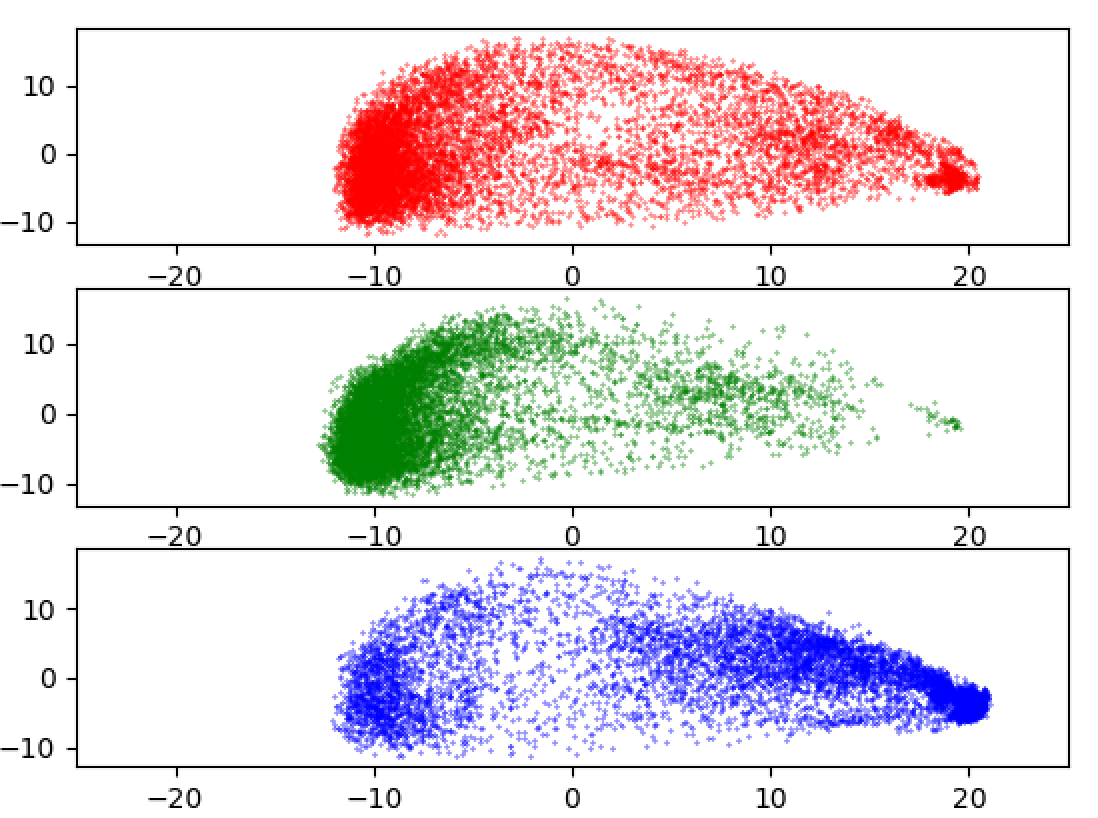} \\ \caption{Output of TDNN3 layer}
\end{subfigure}
\caption{This figure shows the distribution of the various input features across different datasets. The one on the left are the filterbank + pitch features with CMVN used in the baseline experiments, figure in center shows the features that are used as the input to the ASpIRE model (MFCC and i-vector), the right side plots are the domain invariant output of the TDNN3 layer from the ASpIRE model. The red points are \textcolor{Red}{Conversational Speech}, the green ones are \textcolor{ForestGreen}{Broadcast News} and the blue points are from \textcolor{Blue}{Scripted Speech}. We can see that the features become normalized across domains after passing through the ASpIRE model. To visualize this in a 2D space 10000 frames were randomly selected from each of the three datasets and reduced to 2 dimension using Principle Component Analysis (PCA).}
\label{fig:feature_plot}
\end{figure*}

We use CTC-based Sequence Models decoded with WFSTs to train our ASR systems, as detailed in ~\cite{miao2015eesen}. The target labels are generated using a common grapheme-to-phoneme (G2P) library~\cite{mortensen2018epitran} that we use to maintain uniformity of phonemes across datasets, allowing us to do cross-domain experiments. We start this section by briefly explaining the baseline setup and the G2P library, followed by the explanation of the ASpIRE Chain Models~\cite{peddinti2015jhu} (built for the ASpIRE challenge) along with its advantages and how we use it to extract domain invariant features to train our CTC-based models.

\subsection{Baseline Features and Epitran G2P Library}
\textbf{Baseline Input Features}: For the baseline experiments we use the basic features used in EESEN~\cite{miao2015eesen}, 40 dimensional filterbank features along with their first and second order derivatives and 3 pitch features. We also apply mean and variance normalization to the features for each speaker.

\noindent{\textbf{Model Structure}}: As done in ~\cite{dalmia2018sequence} we use a 6 layer Bi-LSTM model with 360 cells in each direction with a window of size 3. Each utterance is sub-sampled with a stride of 2 frames to make 3 equivalent copies of the same utterance. They are trained using the CTC loss function described in \cite{miao2015eesen}.

\noindent{\textbf{Target Labels}}: To maintain uniformity of the target labels across all the Turkish datasets, we use the Epitran~\cite{mortensen2018epitran} grapheme-to-phoneme library to generate lexicons for the words present in the training, development and test set. To verify that the lexicons generated by the Epitran system are reasonable, we compute WER on each of the datasets. A WFST based decoding was performed using a language model that was built on the training data~(lowest perplexity language models was chosen between 3-gram and 4-gram models) and acoustic model output was scaled with the priors of the phonemes in the training data. 
Table~\ref{tab:epitran_wer} shows the Phoneme error rates~(PER) and the corresponding Word error rates~(WER) on baseline models built on Turkish Babel Dataset and Turkish Broadcast news data. 
\begin{table}[H]
\centering
\caption{Phoneme (\% PER) and Word error rate (\% WER) of baseline systems trained on Turkish Broadcast News and Conversational Datasets using the Epitran Phoneme Set.}
\label{tab:epitran_wer}
\begin{tabular}{ccc}
\toprule
Model   										& PER        	& WER     		\\ \midrule
\textcolor{Red}{Conversational Speech}			& 34.5       	& 49.6  		\\
\textcolor{ForestGreen}{Broadcast News}  		& 5.8 			& 20.2			 \\ \bottomrule
\end{tabular}
\end{table}
Going forward, we will analyze the trends in phoneme error rate, which is a good measure of the success of the acoustic-only adaptation performed in this paper. This would also avoid any kind of re-scoring of phoneme sequences done using the language model. 

\subsection{ASpIRE Chain Model}

Originally developed in response to the IARPA ASpIRE challenge~\cite{harper2015automatic},
and trained on augmented English Fisher data~\cite{peddinti2015jhu} using a TDNN architecture~\cite{peddinti2015time}, the ASpIRE model\footnote{The pre-trained model is available at \url{dl.kaldi-asr.org/models/0001_aspire_chain_model.tar.gz}} is robust against
various acoustic environments, including noisy and reverberant domains.


\noindent{\textbf{Proposed Input Features}}: Since the ASpIRE model is supposed to do well in mismatched conditions, following the ideas from~\cite{dalmia2018sequence}, we want to use the internal features of the English trained ASpIRE chain model to rapidly develop a Turkish CTC based ASR and test for its robustness, which has been explained in detail in the following section.

We use the same model structure as the baseline model, except, we keep a window of size 1 to decrease the training time of the model. Further, the effect of the window size on the result is less prominent when using the domain robust feature extraction technique discussed below.


\section{Domain Robust Feature Extraction}

As the previous section suggests, the ASpIRE chain model is robust to noisy and reverberant environments. We expect that features from the pre-trained ASpIRE model can be helpful when testing on unseen domains even under a cross-lingual settings.

To extract the ASpIRE feature from the TDNN layers, we first prepare the input features into a pre-trained ASpIRE model. 40-dimensional Mel-frequency cepstral coefficients (MFCCs) are used as the main input features to the ASpIRE model. Then 100-dimensional i-vector is appended to each frame, which can capture information of both speaker and environment.

The TDNN architecture used in the ASpIRE chain model is a 5 layer sub-sampled TDNN network. TDNN0 layer is directly connected to the input layer which splices frames from $t-1$ to $t+1$ for time step $t$. We write $\{-1, 0, 1\}$ as the context window for this layer. The TDNN1 layer uses one more future frame, and its context window is $\{-1, 0, 1, 2\}$. The subsequent three TDNN layers (TDNN2, 3, 4) share the same context window $\{-3, 0, 3\}$. Finally, the last TDNN layer (TDNN5) uses $\{-6, -3, 0\}$. In total, each frame after TDNN5 layer has 17 frames as its left context and 12 frames as the right context. 

In our experiments, we explore activations of TDNN3 layer as we observed that it achieved the best WER when compared to the other layers in the cross-lingual adaptation setting. We believe this could be possible because shallow layers such as TDNN1 or TDNN2 might not have learned informative acoustic features that are independent of the input domain. On the other hand, the later layers could be too adapted to the target language and the target domain condition and thus not suitable for cross-lingual and domain mismatch scenarios. For example TDNN5 layer has a context of $\{-6, -3, 0\}$ which was added to handle reverberations~\cite{peddinti2015time}.

To investigate robustness of the TDNN3 activations, in Figure~\ref{fig:feature_plot}, we visualize its distribution on various domains of Turkish datasets that we have and compare it qualitatively with the baseline input features. We also visualize the input to the ASpIRE chain model and show how by the 3rd TDNN layer the model learns to normalize the domain mismatch. 

The TDNN3 layer activations have very similar distributions for the three corpus which motivates us towards using it as a robust ``domain invariant'' feature extractor.

\section{Results and Discussions}

\begin{table*}[ht!]
\centering
\caption{Phoneme Error Rates (\% PER) using different datasets. Two features are evaluated here: CMVN filterbank + pitch with window size 3 and cross-lingual ASpIRE features from the TDNN3 layer. Each model is trained with either conversational speech or broadcast news, and then tested using all three datasets to show their ability of adapt to a new environment. The colored corpus in the first row denotes the test dataset, and the second column is the training dataset.}
\label{tab:main-results}
\begin{tabular}{ccccc}
\toprule
Model                               &                         & \textcolor{red}{Conversational}       & \textcolor{ForestGreen}{Broadcast}                 & \textcolor{blue}{Scripted}                     \\
                                    &                         & \textcolor{red}{Speech}               & \textcolor{ForestGreen}{News}                      & \textcolor{blue}{Speech}                   \\ \midrule
\multicolumn{1}{c|}{Baseline} & Conversational Speech   & 34.5                 & 34.8                      & 40.1                     \\
\multicolumn{1}{c|}{CMVN FBank + Pitch Win-3}       & Broadcast News        & 60.8                 & 5.8                       & 67.1                     \\
\midrule \midrule
\multicolumn{1}{c|}{Domain Robust}         & Conversational Speech   & \textbf{32.6}        & \textbf{24.7}             & \textbf{33.2}            \\
\multicolumn{1}{c|}{ASPIRE TDNN3 layer}  & Broadcast News         & \textbf{53.4}        & \textbf{4.9}              & \textbf{35.0}              \\
\bottomrule
\end{tabular}
\end{table*}

\subsection{Proposed Approach}
We investigate the performance of the domain robust ASpIRE features in a model of window size 1 and compare them with systems trained using baseline input features (filter banks). We train separate models on both \textcolor{red}{Conversational Speech} and \textcolor{ForestGreen}{Broadcast News} and evaluate them using the other 2 datasets, along with \textcolor{blue}{Scripted Speech}. This evaluation helps us understand the robustness of our model and its ability to recognize speech of unseen environments. 

Our results are summarized in Table~\ref{tab:main-results}. The table shows phoneme error rate (PER) of various train/test combinations. Here, for example, the upper-right number, 40.1\,\% denotes the PER when the model is trained with Conversational Speech and evaluated with \textcolor{blue}{Scripted Speech}.

We can see that these features help in both in-domain and out-of-domain setting. We see a relative improvement 15.5\% for Broadcast News and 5.5\% for Conversational Speech when testing on the same dataset. This corroborates the in-domain adaptation improvements shown in~\cite{dalmia2018sequence}, and we thus believe that a well trained ASpIRE model is a good starting point for adaptation. 

As expected, the gains on cross-domain experiments are much larger and more significant. We see a relative improvement of 29.0\% PER when testing a Conversational Speech Model on \textcolor{ForestGreen}{Broadcast News} and a relative 47.8\% PER improvement when testing a Broadcast News system on \textcolor{blue}{Scripted Speech}. Other cross-domain systems also show improvements of around 15\% relative. These results show strong indication that using these ASpIRE activations cross-lingually can generalize well to adapt a new low resource language in different domains.




\subsection{Diagnostic Experiments}
To understand the gains that come from the cross-lingual transfer, we performed an experiment by using data augmentation and i-vector components (key-factors towards robustness as mentioned in~\cite{peddinti2015jhu}), and applying them to our baseline model. The results are shown in in Table~\ref{tab:ablation-results}~(notable improvements in bold).

\subsubsection{Data Augmentation}
For this experiment we follow the data augmentation procedure in~\cite{peddinti2015jhu} by creating multi-condition training data by using a collection of 7 datasets, (RWCP~\cite{nakamura2000acoustical}, AIRD~\cite{jeub2009binaural}, Reverb2014~\cite{kinoshita2013reverb}, OpenAIR~\cite{murphy2010openair}, MARDY~\cite{wen2006evaluation}, QMUL~\cite{stowell2013open}, Impulse responses from Varechoic chamber~\cite{neto1999itu}),
to get different real world room impulse responses (RIR) and noise recordings. Three different copies of the original dataset were created with a randomly chosen room impulse response. Noise was added when available with randomly chosen SNR value between 0, 5, 10, 15 and 20 dB. The 3 copies along with the original copy was used to train a model with baseline filterbank features with a window size of 3. As per the results in Table~\ref{tab:ablation-results}, we note that the Broadcast News system tested on \textcolor{blue}{Scripted Speech} improves by 15\% relative PER w.r.t. to baseline system shown in Table~\ref{tab:main-results}. Other cross-domain systems don't show any notable improvements. We believe that this may be because data augmentation is meant to simulate the testing conditions and we did not use any such prior information while creating the augmentations.

\begin{table}[H]
\centering
\caption{Phoneme Error Rates~(\%~PER) on Baseline~(CMVN filterbank+pitch) features with window size of 3 using data augmentation and i-vectors.}
\label{tab:ablation-results}
\begin{tabular}{ccccc}
\toprule
Model                               &                         & \textcolor{red}{Conv.}       & \textcolor{ForestGreen}{Broad.}                 & \textcolor{blue}{Scripted}                     \\
                                    &                         & \textcolor{red}{Speech}               & \textcolor{ForestGreen}{News}                      & \textcolor{blue}{Speech}                   \\ \midrule
\multicolumn{1}{c|}{Data} & Conv. Speech   & 35.8                 & 37.2                      & 41.5                     \\
\multicolumn{1}{c|}{Aug.}       & Broad. News        & 60.0                 & 6.1                      & \textbf{56.9}                     \\
\midrule
\multicolumn{1}{c|}{i-vec.}         & Conv. Speech   & 34.1        & 34.3             & 40.2            \\
\multicolumn{1}{c|}{}  & Broad. News         & 61.0        & 5.7              & 78.7              \\
\bottomrule
\end{tabular}
\end{table}

\subsubsection{i-vectors}
For this experiment, we use the same background model as that of the pre-trained ASpIRE model to extract i-vectors for each sub-speaker~(6000 frames of a single speaker is called a subspeaker) in the dataset, as getting reliable background model and training a good i-vector can be difficult for ``real'' low-resource languages. As per the results in Table~\ref{tab:ablation-results}, we see that i-vectors did not see any notable improvements in performance. However, it had a big drop when testing the Broadcast News system on \textcolor{blue}{Scripted Speech}. We think that this maybe be because of using ASpIRE background model to extract i-vectors and it maybe a better idea to train them from scratch. 

The diagnostics show that techniques like data augmentation and i-vectors need to be prepared carefully and require a lot of tuning in the development data before it actually becomes helpful. Therefore for scenarios like LORELEI, where rapid prototyping is needed, this maybe to difficult to tune.

\subsection{Decoding using a matched Language Model - \\ LORELEI Simulation}

The described approach responds to the requirements of the DARPA LORELEI (Low Resource Languages for Emergent Incidents) program. Using the robust feature extractor, it is possible to rapidly develop a speech recognition system in an ``incident’' language, even if there is significant channel mis-match between available training data and the test data. In past NIST’s LoReHLT evaluations,\footnote{https://www.nist.gov/itl/iad/mig/lorehlt-evaluations} we found that for practical systems, channel and domain mis-match is a significant problem, which is ignored by existing cross- and multi-lingual training work. For the 2018 LoReHLT evaluation, the authors built two speech recognizers (for Kinyarwanda and Sinhala) within about 24\,hours, using ``available’' and minimal (or modest) data resources. A description of these systems will be submitted to a later conference, as soon as analysis of the evaluation results has concluded. On our internal test set, we found that the robust feature extractor described in this paper was similarly effective for the two LoReHLT 2018 languages (where only clean training and development data from few speakers was available, while the test data consists of broadcast material), as for Turkish.
\begin{table}[H]
\centering
\caption{Word Error Rates~(\%~WER) on Broadcast News data for different Conversational Speech models}
\label{tab:lorelei-results}
\begin{tabular}{ccc}
\toprule
Conv. Speech                              			&     \textcolor{ForestGreen}{Broadcast News} WER (\%)        \\ \midrule
Baseline feats   									&     50.4           				\\
ASpIRE TDNN3 feats        							&     \textbf{34.7}                 \\  \bottomrule
\end{tabular}
\end{table}
To get an estimate of the overall improvement in the word error rate, we can simulate a LORELEI-like scenario by using our Conversational Speech model to decode Broadcast News with an in-domain language model. Table~\ref{tab:lorelei-results} shows that the WER goes down considerably~(30\% relative improvement) after using the domain robust ASpIRE TDNN3 layer features.

\section{Conclusions}

In this paper, we present a simple yet effective approach to improve domain robustness in low-resource, cross-lingual ASR. Our target phoneme error rate on Turkish Broadcast News data improved by 10\,\% absolute (almost 30\,\% relative) by porting a robust feature extractor from a well-resourced source language to the target language, and training on mis-matched (Babel FLP) data. With a target-domain language model, word error rate improves similarly by about 15\,\% absolute~(30\,\% relative). Large improvements are also observed for the recognition of scripted speech, independent of training material. The feature extractor does not depend on the target language, so the approach is suitable for rapid prototyping and bootstrapping; we believe it will be even more beneficial for even smaller training corpora (Babel LLP, VLLP).

\section{Acknowledgements}
This project was sponsored by the Defense Advanced Research Projects Agency (DARPA) Information Innovation Office (I2O), program: Low Resource Languages for Emergent Incidents (LORELEI), issued by DARPA/I2O under Contract No. HR0011-15-C-0114. 

We gratefully acknowledge the support of NVIDIA Corporation with the donation of the Titan X Pascal GPU used for this research. The authors would like to thank Anant Subramanian, Soumya Wadhwa and Shivani Poddar for sharing their insights. The authors would also like to thank Murat Saraclar and Ebru Arisoy for their help in establishing the Turkish BN baselines. 
 \bibliographystyle{IEEEbib}
 \bibliography{mybib}

\begin{thebibliography}{10}

\bibitem{ko2015audio}
Tom Ko, Vijayaditya Peddinti, Daniel Povey, and Sanjeev Khudanpur,
\newblock ``Audio augmentation for speech recognition,''
\newblock in {\em Sixteenth Annual Conference of the International Speech
  Communication Association}, 2015.

\bibitem{peddinti2015jhu}
Vijayaditya Peddinti, Guoguo Chen, Vimal Manohar, Tom Ko, Daniel Povey, and
  Sanjeev Khudanpur,
\newblock ``{JHU aspire system: Robust LVSCR with TDNNs, ivector adaptation and
  RNN-LMs},''
\newblock in {\em Automatic Speech Recognition and Understanding (ASRU), 2015
  IEEE Workshop on}. IEEE, 2015, pp. 539--546.

\bibitem{lorelei}
``{Low Resource Languages for Emergent Incidents (LORELEI)},
  howpublished={\url{https://www.darpa.mil/program/low-resource-languages-for-emergent-incidents}},''
  .

\bibitem{barker2015third}
Jon Barker, Ricard Marxer, Emmanuel Vincent, and Shinji Watanabe,
\newblock ``{The third ‘CHiME’speech separation and recognition challenge:
  Dataset, task and baselines},''
\newblock in {\em Automatic Speech Recognition and Understanding (ASRU), 2015
  IEEE Workshop on}. IEEE, 2015, pp. 504--511.

\bibitem{harper2015automatic}
Mary Harper,
\newblock ``{The automatic speech recogition in reverberant environments
  (ASpIRE) challenge},''
\newblock in {\em Automatic Speech Recognition and Understanding (ASRU), 2015
  IEEE Workshop on}. IEEE, 2015, pp. 547--554.

\bibitem{kinoshita2016summary}
Keisuke Kinoshita, Marc Delcroix, Sharon Gannot, Emanu{\"e}l~AP Habets,
  Reinhold Haeb-Umbach, Walter Kellermann, Volker Leutnant, Roland Maas,
  Tomohiro Nakatani, Bhiksha Raj, et~al.,
\newblock ``A summary of the {REVERB} challenge: state-of-the-art and remaining
  challenges in reverberant speech processing research,''
\newblock {\em EURASIP Journal on Advances in Signal Processing}, vol. 2016,
  no. 1, pp. 7, 2016.

\bibitem{mitra2016coping}
Vikramjit Mitra and Horacio Franco,
\newblock ``Coping with unseen data conditions: Investigating neural net
  architectures, robust features, and information fusion for robust speech
  recognition.,''
\newblock in {\em INTERSPEECH}, 2016, pp. 3783--3787.

\bibitem{adiga2013gammatone}
Aniruddha Adiga, Mathew Magimai, and Chandra~Sekhar Seelamantula,
\newblock ``Gammatone wavelet cepstral coefficients for robust speech
  recognition,''
\newblock in {\em TENCON 2013-2013 IEEE Region 10 Conference (31194)}. IEEE,
  2013, pp. 1--4.

\bibitem{mitra2012normalized}
Vikramjit Mitra, Horacio Franco, Martin Graciarena, and Arindam Mandal,
\newblock ``Normalized amplitude modulation features for large vocabulary
  noise-robust speech recognition,''
\newblock in {\em Acoustics, Speech and Signal Processing (ICASSP), 2012 IEEE
  International Conference on}. IEEE, 2012, pp. 4117--4120.

\bibitem{mitra2013damped}
Vikramjit Mitra, Horacio Franco, and Martin Graciarena,
\newblock ``Damped oscillator cepstral coefficients for robust speech
  recognition.,''
\newblock in {\em Interspeech}, 2013, pp. 886--890.

\bibitem{gehring2013extracting}
Jonas Gehring, Yajie Miao, Florian Metze, and Alex Waibel,
\newblock ``Extracting deep bottleneck features using stacked auto-encoders,''
\newblock in {\em Acoustics, Speech and Signal Processing (ICASSP), 2013 IEEE
  International Conference on}. IEEE, 2013, pp. 3377--3381.

\bibitem{yu2011improved}
Dong Yu and Michael~L Seltzer,
\newblock ``Improved bottleneck features using pretrained deep neural
  networks,''
\newblock in {\em Twelfth Annual Conference of the International Speech
  Communication Association}, 2011.

\bibitem{karafiat2015three}
Martin Karafi{\'a}t, Franti{\v{s}}ek Gr{\'e}zl, Luk{\'a}{\v{s}} Burget, Igor
  Sz{\"o}ke, and Jan {\v{C}}ernock{\`y},
\newblock ``Three ways to adapt a {CTS} recognizer to unseen reverberated
  speech in {BUT} system for the {ASpIRE} challenge,''
\newblock in {\em Sixteenth Annual Conference of the International Speech
  Communication Association}, 2015.

\bibitem{hsiao2015robust}
Roger Hsiao, Jeff Ma, William Hartmann, Martin Karafi{\'a}t, Franti{\v{s}}ek
  Gr{\'e}zl, Luk{\'a}{\v{s}} Burget, Igor Sz{\"o}ke, Jan~Honza
  {\v{C}}ernock{\`y}, Shinji Watanabe, Zhuo Chen, et~al.,
\newblock ``Robust speech recognition in unknown reverberant and noisy
  conditions,''
\newblock in {\em Automatic Speech Recognition and Understanding (ASRU), 2015
  IEEE Workshop on}. IEEE, 2015, pp. 533--538.

\bibitem{swietojanski2014learning}
Pawel Swietojanski and Steve Renals,
\newblock ``Learning hidden unit contributions for unsupervised speaker
  adaptation of neural network acoustic models,''
\newblock in {\em Spoken Language Technology Workshop (SLT), 2014 IEEE}. IEEE,
  2014, pp. 171--176.

\bibitem{swietojanski2015differentiable}
Pawel Swietojanski and Steve Renals,
\newblock ``Differentiable pooling for unsupervised speaker adaptation,''
\newblock in {\em Acoustics, Speech and Signal Processing (ICASSP), 2015 IEEE
  International Conference on}. IEEE, 2015, pp. 4305--4309.

\bibitem{meng2017unsupervised}
Zhong Meng, Zhuo Chen, Vadim Mazalov, Jinyu Li, and Yifan Gong,
\newblock ``Unsupervised adaptation with domain separation networks for robust
  speech recognition,''
\newblock {\em arXiv preprint arXiv:1711.08010}, 2017.

\bibitem{sun2017unsupervised}
Sining Sun, Binbin Zhang, Lei Xie, and Yanning Zhang,
\newblock ``An unsupervised deep domain adaptation approach for robust speech
  recognition,''
\newblock {\em Neurocomputing}, vol. 257, pp. 79--87, 2017.

\bibitem{shinohara2016adversarial}
Yusuke Shinohara,
\newblock ``Adversarial multi-task learning of deep neural networks for robust
  speech recognition.,''
\newblock {\em Proc. Interspeech 2016}, 2016.

\bibitem{schultz1997fast}
Tanja Schultz and Alex Waibel,
\newblock ``Fast bootstrapping of {LVCSR} systems with multilingual phoneme
  sets,''
\newblock in {\em Fifth European Conference on Speech Communication and
  Technology}, 1997.

\bibitem{metze2015semi}
Florian Metze, Ankur Gandhe, Yajie Miao, Zaid Sheikh, Yun Wang, Di~Xu, Hao
  Zhang, Jungsuk Kim, Ian Lane, Won~Kyum Lee, et~al.,
\newblock ``Semi-supervised training in low-resource {ASR} and {KWS},''
\newblock in {\em Acoustics, Speech and Signal Processing (ICASSP), 2015 IEEE
  International Conference on}. IEEE, 2015, pp. 4699--4703.

\bibitem{xue2014fast}
Shaofei Xue, Ossama Abdel-Hamid, Hui Jiang, Lirong Dai, and Qingfeng Liu,
\newblock ``Fast adaptation of deep neural network based on discriminant codes
  for speech recognition,''
\newblock {\em IEEE/ACM Transactions on Audio, Speech, and Language
  Processing}, vol. 22, no. 12, pp. 1713--1725, 2014.

\bibitem{vu2012multilingual}
Ngoc~Thang Vu, Florian Metze, and Tanja Schultz,
\newblock ``Multilingual bottle-neck features and its application for
  under-resourced languages,''
\newblock in {\em Spoken Language Technologies for Under-Resourced Languages},
  2012.

\bibitem{thomas2012multilingual}
Samuel Thomas, Sriram Ganapathy, and Hynek Hermansky,
\newblock ``Multilingual {MLP} features for low-resource {LVCSR} systems,''
\newblock in {\em Acoustics, Speech and Signal Processing (ICASSP), 2012 IEEE
  International Conference on}. IEEE, 2012, pp. 4269--4272.

\bibitem{knill2013investigation}
Kate~M Knill, Mark~JF Gales, Shakti~P Rath, Philip~C Woodland, Chao Zhang, and
  S-X Zhang,
\newblock ``Investigation of multilingual deep neural networks for spoken term
  detection,''
\newblock in {\em Automatic Speech Recognition and Understanding (ASRU), 2013
  IEEE Workshop on}. IEEE, 2013, pp. 138--143.

\bibitem{grezl2014adaptation}
Frantisek Gr{\'e}zl, Martin Karafi{\'a}t, and Karel Vesely,
\newblock ``Adaptation of multilingual stacked bottle-neck neural network
  structure for new language,''
\newblock in {\em Acoustics, Speech and Signal Processing (ICASSP), 2014 IEEE
  International Conference on}. IEEE, 2014, pp. 7654--7658.

\bibitem{miao2014distributed}
Yajie Miao, Hao Zhang, and Florian Metze,
\newblock ``Distributed learning of multilingual {DNN} feature extractors using
  {GPUs},''
\newblock in {\em Fifteenth Annual Conference of the International Speech
  Communication Association}, 2014.

\bibitem{huang2013cross}
Jui-Ting Huang, Jinyu Li, Dong Yu, Li~Deng, and Yifan Gong,
\newblock ``Cross-language knowledge transfer using multilingual deep neural
  network with shared hidden layers,''
\newblock in {\em Acoustics, Speech and Signal Processing (ICASSP), 2013 IEEE
  International Conference on}. IEEE, 2013, pp. 7304--7308.

\bibitem{ghoshal2013multilingual}
Arnab Ghoshal, Pawel Swietojanski, and Steve Renals,
\newblock ``Multilingual training of deep neural networks,''
\newblock in {\em Acoustics, Speech and Signal Processing (ICASSP), 2013 IEEE
  International Conference on}. IEEE, 2013, pp. 7319--7323.

\bibitem{stolcke2006cross}
Andreas Stolcke, Frantisek Grezl, Mei-Yuh Hwang, Xin Lei, Nelson Morgan, and
  Dimitra Vergyri,
\newblock ``Cross-domain and cross-language portability of acoustic features
  estimated by multilayer perceptrons,''
\newblock in {\em Acoustics, Speech and Signal Processing, 2006. ICASSP 2006
  Proceedings. 2006 IEEE International Conference on}. IEEE, 2006, vol.~1, pp.
  I--I.

\bibitem{dehak2011front}
Najim Dehak, Patrick~J Kenny, R{\'e}da Dehak, Pierre Dumouchel, and Pierre
  Ouellet,
\newblock ``Front-end factor analysis for speaker verification,''
\newblock {\em IEEE Transactions on Audio, Speech, and Language Processing},
  vol. 19, no. 4, pp. 788--798, 2011.

\bibitem{miao2015speaker}
Yajie Miao, Hao Zhang, and Florian Metze,
\newblock ``Speaker adaptive training of deep neural network acoustic models
  using i-vectors,''
\newblock {\em IEEE/ACM Transactions on Audio, Speech and Language Processing
  (TASLP)}, vol. 23, no. 11, pp. 1938--1949, 2015.

\bibitem{dalmia2018sequence}
Siddharth Dalmia, Ramon Sanabria, Florian Metze, and Alan~W Black,
\newblock ``Sequence-based multi-lingual low resource speech recognition,''
\newblock in {\em 2018 IEEE International Conference on Acoustics, Speech and
  Signal Processing (ICASSP)}. IEEE, 2018, pp. 4909--4913.

\bibitem{tongfast}
Sibo Tong, Philip~N Garner, and Herv{\'e} Bourlard,
\newblock ``Fast language adaptation using phonological information,''
\newblock {\em Proc. Interspeech 2018}, pp. 2459--2463, 2018.

\bibitem{arisoy2009turkish}
Ebru Arisoy, Dogan Can, Siddika Parlak, Hasim Sak, and Murat Sara{\c{c}}lar,
\newblock ``Turkish broadcast news transcription and retrieval,''
\newblock {\em IEEE Transactions on Audio, Speech, and Language Processing},
  vol. 17, no. 5, pp. 874--883, 2009.

\bibitem{miao2015eesen}
Yajie Miao, Mohammad Gowayyed, and Florian Metze,
\newblock ``{EESEN}: End-to-end speech recognition using deep {RNN} models and
  {WFST}-based decoding,''
\newblock in {\em Automatic Speech Recognition and Understanding (ASRU), 2015
  IEEE Workshop on}. IEEE, 2015, pp. 167--174.

\bibitem{mortensen2018epitran}
David~R. Mortensen, Siddharth Dalmia, and Patrick Littell,
\newblock ``Epitran: Precision {G2P} for many languages,''
\newblock in {\em Proceedings of the Eleventh International Conference on
  Language Resources and Evaluation (LREC 2018)}, May 2018.

\bibitem{peddinti2015time}
Vijayaditya Peddinti, Daniel Povey, and Sanjeev Khudanpur,
\newblock ``A time delay neural network architecture for efficient modeling of
  long temporal contexts,''
\newblock in {\em Sixteenth Annual Conference of the International Speech
  Communication Association}, 2015.

\bibitem{nakamura2000acoustical}
Satoshi Nakamura, Kazuo Hiyane, Futoshi Asano, Takanobu Nishiura, and Takeshi
  Yamada,
\newblock ``Acoustical sound database in real environments for sound scene
  understanding and hands-free speech recognition.,''
\newblock in {\em LREC}, 2000.

\bibitem{jeub2009binaural}
Marco Jeub, Magnus Schafer, and Peter Vary,
\newblock ``A binaural room impulse response database for the evaluation of
  dereverberation algorithms,''
\newblock in {\em Digital Signal Processing, 2009 16th International Conference
  on}. IEEE, 2009, pp. 1--5.

\bibitem{kinoshita2013reverb}
Keisuke Kinoshita, Marc Delcroix, Takuya Yoshioka, Tomohiro Nakatani, Armin
  Sehr, Walter Kellermann, and Roland Maas,
\newblock ``The {REVERB} challenge: A common evaluation framework for
  dereverberation and recognition of reverberant speech,''
\newblock in {\em Applications of Signal Processing to Audio and Acoustics
  (WASPAA), 2013 IEEE Workshop on}. IEEE, 2013, pp. 1--4.

\bibitem{murphy2010openair}
Damian~T Murphy and Simon Shelley,
\newblock ``Openair: An interactive auralization web resource and database,''
\newblock in {\em Audio Engineering Society Convention 129}. Audio Engineering
  Society, 2010.

\bibitem{wen2006evaluation}
Jimi~YC Wen, Nikolay~D Gaubitch, Emanuel~AP Habets, Tony Myatt, and Patrick~A
  Naylor,
\newblock ``Evaluation of speech dereverberation algorithms using the {MARDY}
  database,''
\newblock in {\em in Proc. Intl. Workshop Acoust. Echo Noise Control (IWAENC}.
  Citeseer, 2006.

\bibitem{stowell2013open}
Dan Stowell and Mark~D Plumbley,
\newblock ``An open dataset for research on audio field recording archives:
  freefield1010,''
\newblock {\em arXiv preprint arXiv:1309.5275}, 2013.

\bibitem{neto1999itu}
Simao Ferraz De~Campos Neto,
\newblock ``The {ITU-T} software tool library,''
\newblock {\em International journal of speech technology}, vol. 2, no. 4, pp.
  259--272, 1999.

\end{thebibliography}
\end{document}